\documentclass[10pt,twocolumn,letterpaper]{article}

\usepackage{wacv}              

\usepackage{graphicx}
\usepackage{amsmath}
\usepackage{amssymb}
\usepackage{booktabs}
\usepackage{xcolor}

\usepackage{multirow}
\usepackage{array}
\usepackage{siunitx}
\usepackage{makecell} 
\usepackage{wrapfig}
\usepackage{subcaption}
\usepackage{caption}
\usepackage{comment}
\usepackage{bbding}

\usepackage{arydshln}
\usepackage{float}

\def\method{WiGNet}

\newcommand\blfootnote[1]{%
  \begingroup
  \renewcommand\thefootnote{}\footnote{#1}%
  \addtocounter{footnote}{-1}%
  \endgroup
}

\definecolor{wignet}{HTML}{b85450}
\definecolor{greedyvig}{HTML}{82b366}
\definecolor{mobilevig}{HTML}{d6b656}
\definecolor{vig}{HTML}{6c8ebf}
\definecolor{resnet}{HTML}{FF77FF}
\definecolor{transformer}{HTML}{7FFFD0}
\definecolor{mlp}{HTML}{3B444B}

\usepackage[pagebackref,breaklinks,colorlinks]{hyperref}

\usepackage[capitalize]{cleveref}
\crefname{section}{Sec.}{Secs.}
\Crefname{section}{Section}{Sections}
\Crefname{table}{Table}{Tables}
\crefname{table}{Tab.}{Tabs.}

\def\wacvPaperID{626} 
\def\confName{WACV}
\def\confYear{2025}

\begin{document}

\title{WiGNet: Windowed Vision Graph Neural Network}

\author{ 
\begin{tabular}{c}
Gabriele Spadaro$^{1,2}$ \quad Marco Grangetto$^{1}$ \quad Attilio Fiandrotti$^{1,2}$ \quad Enzo Tartaglione$^{2}$ \quad Jhony H. Giraldo$^{2}$
\end{tabular}
\\
$^1$University of Turin, Italy  ~~ \\ $^2$ LTCI, T\'el\'ecom Paris, Institut Polytechnique de Paris~~ \\ 
 {\tt\small gabriele.spadaro@unito.it}
}
\maketitle

\begin{abstract}
In recent years, Graph Neural Networks (GNNs) have demonstrated strong adaptability to various real-world challenges, with architectures such as Vision GNN (ViG) achieving state-of-the-art performance in several computer vision tasks.
However, their practical applicability is hindered by the computational complexity of constructing the graph, which scales quadratically with the image size.
In this paper, we introduce a novel \textbf{Wi}ndowed vision \textbf{G}raph neural \textbf{Net}work (\method) model for efficient image processing.
\method~explores a different strategy from previous works by partitioning the image into windows and constructing a graph within each window.
Therefore, our model uses graph convolutions instead of the typical 2D convolution or self-attention mechanism.
\method~effectively manages computational and memory complexity for large image sizes.
We evaluate our method in the ImageNet-1k benchmark dataset and test the adaptability of \method~using the CelebA-HQ dataset as a downstream task with higher-resolution images. In both of these scenarios, our method achieves competitive results compared to previous vision GNNs while keeping memory and computational complexity at bay.
\method~offers a promising solution toward the deployment of vision GNNs in real-world applications.
We publicly released the code at \href{https://github.com/EIDOSLAB/WiGNet}{https://github.com/EIDOSLAB/WiGNet}.
\blfootnote{This article has been accepted for publication at the 2025 IEEE/CVF Winter Conference on Applications of Computer Vision (WACV 2025).}

\end{abstract}

\begin{figure}[t]
    \centering
    \includegraphics[width=.9\columnwidth]{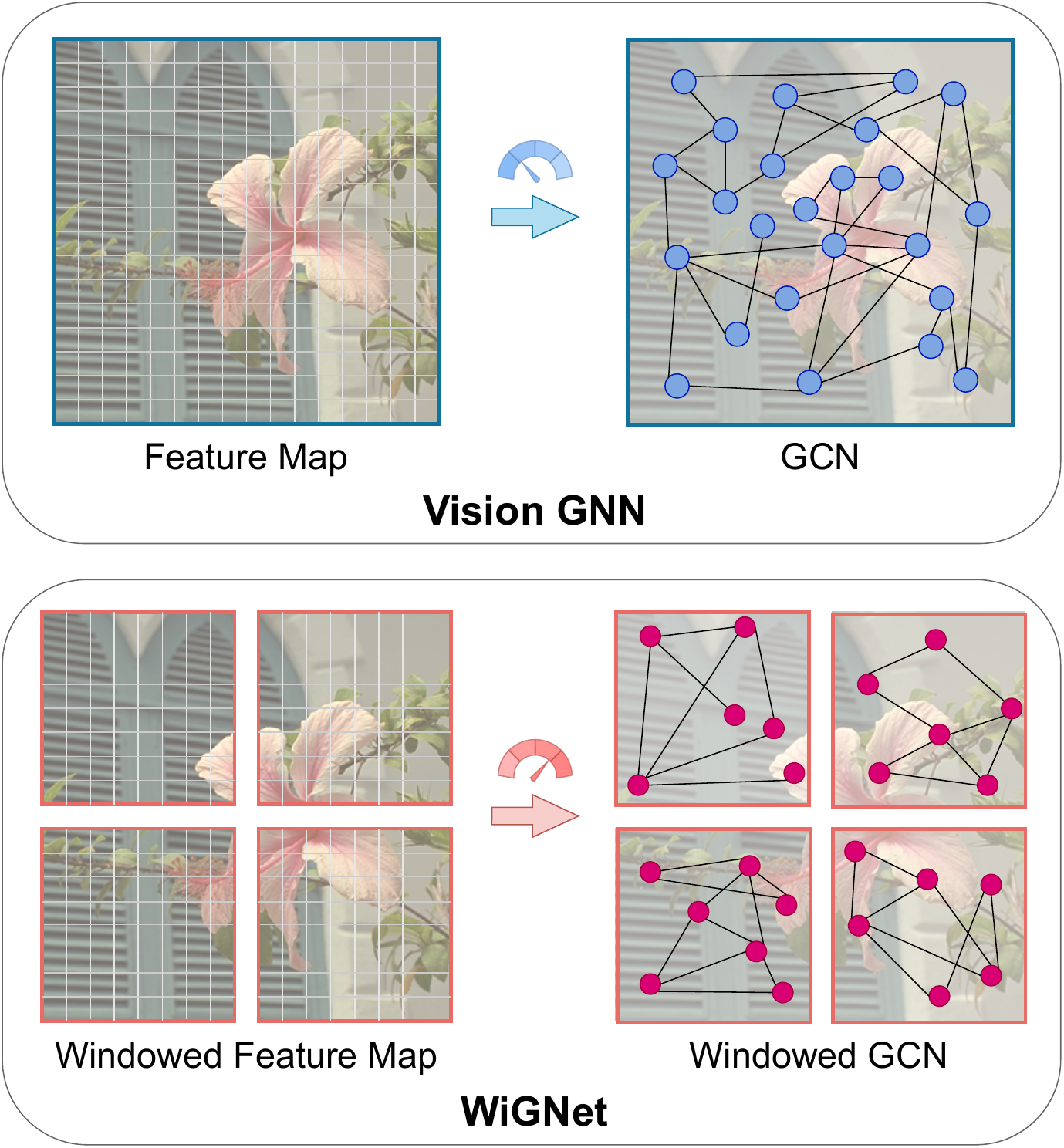}
    \caption{Implementation of Vision GNN (ViG) \cite{vgnn} and the proposed \method. Our model first divides images into local windows where graphs are built. This fundamental change dramatically increases computational and memory efficiency in vision tasks.}
    \label{fig:teaser}
\end{figure}
\section{Introduction}
\label{sec:introduction}

In the last decade, the field of computer vision has progressed significantly, largely due to the success of deep neural networks~\cite{lecun2015deep}.
These models are now established as state-of-the-art in several tasks such as image classification, object detection, semantic segmentation, etc \cite{resnet, fastrcnn, redmon2016you, unet, transfdet, balle17}.
In particular, Convolutional Neural Networks (CNNs)~\cite{lecun1998gradient} exploit the locality of natural images to extract features while Vision Transformers (ViTs)~\cite{vit,swin,deit} implement the attention operator to exploit long-range dependencies of the input image.
Recently, vision-based Graph Neural Networks (GNNs)~\cite{vgnn,mobilevig,greedyvig}, have been proposed with promising results for vision tasks.
Vision GNNs first build a graph over the features extracted from the image, and then apply graph convolutions instead of regular 2D convolutions (CNNs) or the self-attention mechanism (ViTs).
As a result, vision GNNs have benefited from the rich literature of GNNs~\cite{wu2020comprehensive}, adding a new dimension to the landscape of deep learning for image analysis.

Despite the promising results achieved by vision GNNs, there still remain some open challenges.
More precisely, the Vision GNN (ViG) model~\cite{vgnn} relies on the $k$-Nearest Neighbors ($k$-NN) method to construct the graph.
Therefore, the computational complexity of ViG increases quadratically with the number of nodes (patches) extracted from the image and hence with the image size (please refer to Fig. \ref{fig:computational_complexity} for further details).
This hinders their applicability to real-world large-scale datasets and high-resolution images, limiting their practical use.
Overcoming the scalability issue in vision GNNs is crucial for their wider adoption and deployment in real-world applications.

To address the scalability challenges of vision GNNs, we propose a novel \textbf{Wi}ndowed vision \textbf{G}raph neural \textbf{Net}work (\method) model. Similar to previous vision GNNs, \method~treats an image as a graph, yet in a fundamentally different manner.
Namely, the image is first partitioned into non-overlapping windows and only then a separate graph is built for each window as shown in Fig.~\ref{fig:teaser}.
The complexity of building the graphs in our approach grows only linearly with the number of windows,
while maintaining competitive results in image classification tasks.
By focusing on localized regions within the image through windowed processing, \method~efficiently captures relevant features while alleviating the scalability issues encountered by previous vision GNN approaches.

Our work makes the following significant contributions:
\begin{itemize}
    \itemsep0em 
    \item To the best of our knowledge, we are the first to introduce the concept of windowed processing in the context of vision GNNs.
    \item We show that the computational and memory complexity of \method~only grows linearly with the image size, paving the way for broader applications of graph-based models in computer vision.
    \item We thoroughly validate \method~in the ImageNet-1k benchmark dataset and test its adaptability as a feature extractor on CelebA-HQ as a downstream task with higher resolution images. In both of these scenarios, \method~outperforms or obtains competitive performance regarding previous deep learning models like CNNs, ViTs, and ViGs.
\end{itemize}
Our results in ImageNet-1k suggest that \method~successfully exploits vision GNNs for image classification tasks. In addition, classification results of higher-resolution images show that \method~is able to achieve state-of-the-art results while keeping complexity under control.

\section{Related Work}


In this section, we first give an overview of deep learning networks typically adopted in computer vision like CNNs and ViTs.
Then we review GNNs, analyzing their applications to visual tasks and their limitations.

\subsection{CNNs and ViTs.}

Convolutional Neural Networks (CNNs) started dominating the computer vision field since the seminal AlexNet~\cite{alexnet} paper.
CNNs exploit the locality of pixels to extract features from the input image useful to the task on which they are trained. CNNs represented the de-facto standard to solve different tasks from image classification to object detection \cite{redmon2016you, fastrcnn}, semantic segmentation \cite{maskrcnn, unet}, image compression \cite{balle17, balle18}, and many others. 
The rapid development that these architectures have experienced over the past decade has led to the development of models such as ResNet~\cite{resnet} and MobileNet~\cite{mobilenet}, among others \cite{densenet, efficientnet, googlenet}.

More recently, researchers in computer vision have focused on Visual Transformers (ViTs), with the self-attention mechanism at its core.
ViTs build upon the attention mechanism proposed by the Transformer architecture~\cite{transformer} for Natural Language Processing (NLP) tasks in origin, and later on applied with success to different computer vision tasks \cite{nonlocalnn,transfdet, vit, swin}.
Attention enables capturing long-range dependencies between pixels, achieving state-of-the-art results in several computer vision tasks.
The Swin Transformer \cite{swin}, in particular, proposed a hierarchical Transformer-based architecture to extract tokens at different scales and work with high-resolution images.
The multi-head self-attention operator in Swin Transformer is computed in non-overlapped windows. To introduce cross-window connections, the authors proposed a shifted window partitioning approach that alternates with the regular window partitioning in consecutive blocks of Swin Transformer. This method allows for connections between neighboring windows in the previous layer, leading to improvements in image classification, object detection, and semantic segmentation \cite{swin}.

The core function in Swin Transformers can be thought of as an attention operator applied in fully connected graphs from windows in images.
\method~is, in essence, different from Swin Transformers since (i) we operate in $k$-NN graphs instead of fully connected graphs, and (ii) we use a GNN function instead of the self-attention mechanism.
These two changes achieve competitive results regarding the Swin Transformer, other ViTs, and CNN models.

\subsection{Graphs in Computer Vision.}
GNNs emerged as an extension of the convolution operation of CNNs for regular-structured data such as images to the graph domain.
GNNs are typically used for learning graph-structured data representations.
Bruna~\etal~\cite{bruna2014spectral} proposed the first modern GNN by extending the convolutional operator of CNNs to graphs.
Incorporating concepts of signal processing on graphs, Defferrard~\etal~\cite{defferrard2016convolutional} introduced localized spectral filtering for graphs.
Later, Kipf~and~Welling~\cite{kipf2017semi} approximated the spectral filtering operation to obtain efficient Graph Convolutional Networks (GCNs).
Inspired by these works, Veličković~\etal~\cite{velickovic2018graph} presented the attention mechanism on GNNs, resulting in a graph where the connection weights are unique and learned for each edge. 
This allows GATs to effectively model complex relationships between nodes, although with increased computational complexity.

Even though GNNs have generally been adopted for graph-based data~\cite{kipf2017semi}, they have recently demonstrated remarkable success when applied to tasks such as image classification~\cite{vgnn, mobilevig, greedyvig} and segmentation~\cite{giraldo2022hypergraph}.
The Vision GNN (ViG) model~\cite{vgnn}, in particular, drew inspiration from the partition concept introduced in ViTs~\cite{vit}, dividing the input image into smaller patches, and considering each of these patches as a node in the graph of the image.
To establish connections between these nodes, the $k$-NN algorithm is adopted by considering the similarity of nodes in the feature space. 
These features 
are updated using graph convolution operators, considering the features of the node itself and those of its neighbors~\cite{gilmer2017neural}.
Following a similar paradigm as Transformers, these features contribute to the classification of the entire graph, thereby classifying the entire image. 


Using a graph can be beneficial in image processing tasks as it allows for the exploitation of non-local dependencies without the need for multiple convolutional layers and to model complex objects having irregular shapes.
Furthermore, graphs are a more general data structure with respect to a grid of pixels (as modeled by CNNs) or a fully connected graph of patches (as modeled by ViTs).
These advantages have led graph-based models to reach state-of-the-art not only in image classification but also in object detection and instance segmentation \cite{vgnn,greedyvig}.
However, vision GNNs still have a very high computational complexity, especially when working with high-resolution images. For this reason, in the first layers of ViG, the graph is constructed in a bipartite way, connecting patches of the original feature maps with a subsample version of it, obtained using a non-learnable subsampling filter. 
Although this approach decreases the complexity of ViG, it still has a quadratic time complexity, making it slow for processing large images.

More recently, new techniques have emerged to reduce the complexity of ViG by focusing on the graph construction phase. MobileViG~\cite{mobilevig}, for instance, proposed a Sparse Vision Graph Attention (SVGA) module, in which the graph is statically constructed, thus without adopting $k$-NN. Here a patch of the image is connected to patches at a certain hop distance on the same row and column. In this way the number of connections depends on the size of the image, rapidly increasing the memory required to perform graph convolutions. Thus, to obtain a mobile-friendly model, MobileViG only adopts the SVGA module in the last stage of the architecture (where the input tensor is smaller) while the previous stages are implemented using classical depth-wise 2D convolutions. This results in a hybrid model in which the CNNs and GNNs are both adopted.
GreedyViG~\cite{greedyvig} proposed a dynamic version of SVDA named Dynamic Axial Graph
Construction (DAGC). This module starts from the same fixed graph of SVDA and dynamically masks some connections. To create this mask, GreedyViG estimates the mean and standard deviation of the Euclidean distances between the patches in the original image and a diagonally flipped version. Moreover, this DAGC module is implemented in each stage of the architecture, making this CNN-GNN model highly memory-intensive.


Unlike these previous methods, \method~always works on the original feature maps and not on an undersampled version (like ViG). Our method partitions these feature maps into windows of fixed size in which the graph can be constructed. Moreover, the $k$-NN operator is not replaced with a fixed graph structure, but we exploit the locality of pixels to reduce the complexity of this operation, making it linear with respect to the size of the input image.

\section{Windowed Vision Graph Neural Network}
\label{sec:method}

This section describes in detail the proposed WiGNet architecture.
Firstly, we provide some background on graphs and GNNs.
Secondly, we describe the architectural design motivating our choices.
Then we discuss the implication of computational complexity, comparing \method~with the ViG model.
Finally, we propose three different WiGNet versions that we use for our experiments in Section \ref{sec:experiments}.

\subsection{Preliminaries}

\noindent \textbf{Graph.}
A graph is a mathematical entity that can be represented as $G=(\mathcal{V},\mathcal{E})$, where $\mathcal{V}=\{1,\dots,N\}$ is the set of nodes, and ${\mathcal{E}\subseteq \{(i,j)\mid i,j\in \mathcal{V}\;{\textrm {and}}\;i\neq j\}}$ is the set of edges between nodes $i$ and $j$.
We can associate $F$-dimensional feature vectors to every $i$-th node in $G$ such that $\mathbf{x}_{i} \in \mathbb{R}^{F}$.
Therefore, we represent the whole set of features in $G$ with the matrix $\mathbf{X} = [\mathbf{x}_1,\mathbf{x}_2,\dots,\mathbf{x}_N]^\top \in \mathbb{R}^{N \times F}$.

\noindent
\textbf{Message passing function.}
In GNNs, the message-passing function is the standard paradigm for computing graph convolutions \cite{gilmer2017neural}.
Let $\mathbf{x}_{i}'$ be the output of a generic graph convolution, we can thus define the message-passing function as follows:
\begin{equation}
    \mathbf{x}_{i}' = \operatorname{UPDATE}\left(\mathbf{x}_{i},\operatorname{AGG}(\{\mathbf{x}_{j}~\forall~j \in \mathcal{N}_i \})\right),
    \label{eqn:update}
\end{equation}
where $\mathcal{N}_i$ is the set of neighbors of $i$, $\operatorname{AGG}(\cdot)$ is a generic function used to aggregate neighbor information, and $\operatorname{UPDATE}(\cdot)$ updates the representation of the node itself.
A graph convolutional layer can be thought of as an implementation of this message-passing operator by concretely defining the update and aggregation functions.



\subsection{WiGNet Architecture}

\begin{figure}[t]
    \centering
    \includegraphics[width=0.9\linewidth]{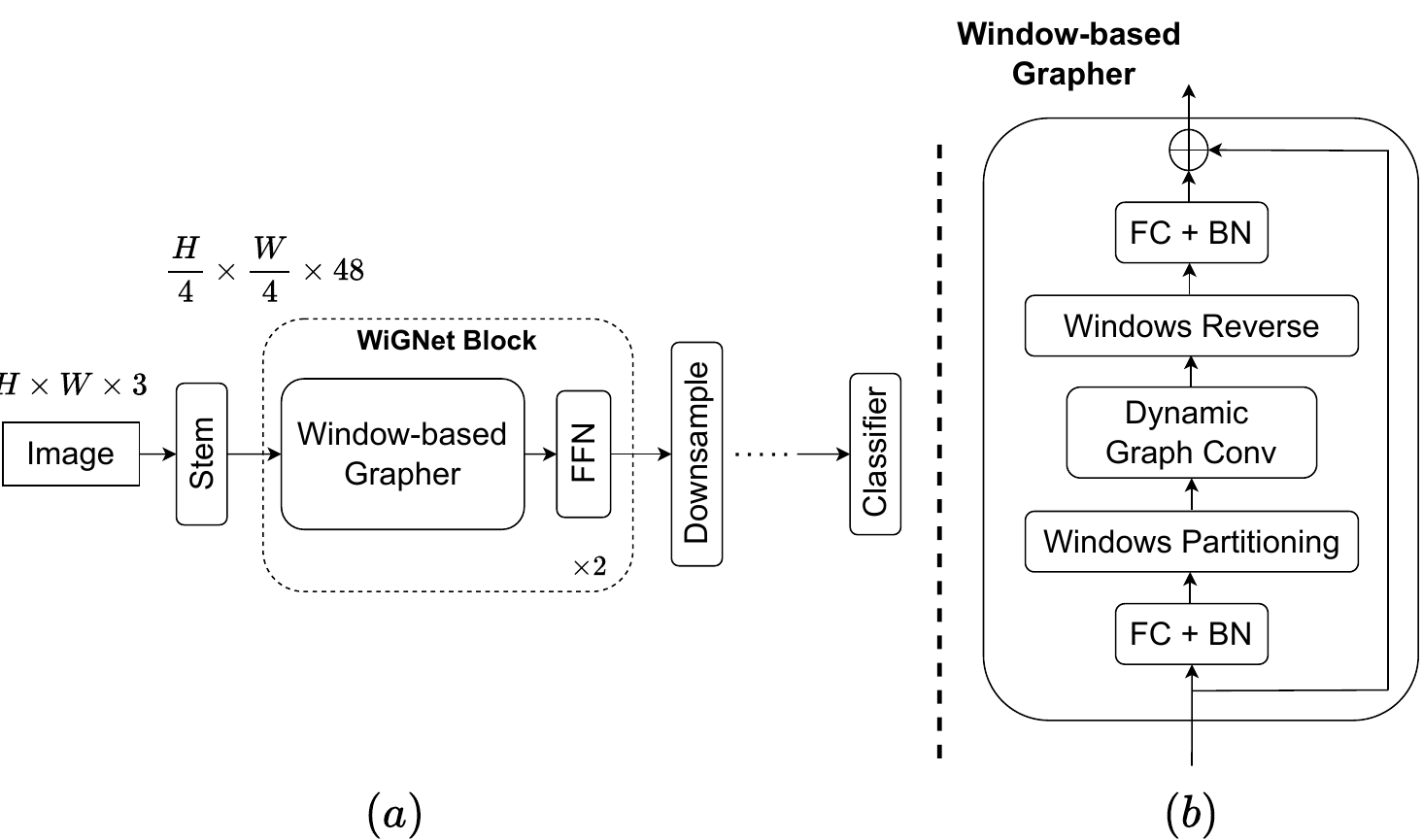}
     \caption{(a): WiGNet architecture exemplified for the \textit{Tiny} version (see Table \ref{tab:wignn} for details). In this example, a linear classifier generates class scores. (b): A graphical illustration of the Window-based Grapher module.}
    \label{fig:arch-grapher}
\end{figure}

\noindent
\textbf{Architecture overview.}
Fig.~\ref{fig:arch-grapher}a shows a bird's eye view of the WiGNet architecture, implementing a four-stage pyramidal feature extractor, where at each stage features of increasingly smaller sizes are extracted. 
A WiGNet is composed of three basic building blocks: (i) the \textit{Stem}, (ii) the \textit{WiGNet block}, and (iii) the \textit{downsampling} module.

The Stem block is a simple feature extractor composed of three convolutional layers that receive as input an image of size $H \times W \times 3$, divides the image into $N$ patches and transforms them into a feature vector $\mathbf{x}_{i} \in \mathbb{R}^{F}$ for each patch, obtaining $\mathbf{X} = [ \mathbf{x}_{1}, \mathbf{x}_{2}, \cdots , \mathbf{x}_{N} ]^\top$.

The WiGNet block is composed of a \textit{Window-based Grapher} module and a \textit{Feed Forward Network} (FFN).
The Grapher module partitions the image into non-overlapping windows, builds a graph for each window, and then local GNN updates are applied to each window.
This is a fundamentally different approach than ViG~\cite{vgnn} where a large graph is built on top of the entire image, with the complexity implications discussed above.
The FFN module further encourages feature diversity. The downsampling block reduces the feature dimension by merging node representations.
Each of the downsampling modules reduces the number of nodes by a factor of $2$ while increasing the size of the feature vectors associated with the remaining nodes.

The complete network architecture is composed of a stack of the Stem function and four WiGNet-plus-Downsampling blocks.
Fig. \ref{fig:arch-grapher}a shows a final fully connected layer for producing class scores.
We propose three versions of WiGNet in Table~\ref{tab:wignn}: (i) tiny (\method-Ti), (ii) small (\method-S), and (iii) medium (\method-M).
In the following, we describe in detail the Grapher module, at the core of WiGNet.

\begin{table}[tb]
	\small
	\centering

	\setlength{\tabcolsep}{3pt}

\resizebox{0.97\columnwidth}{!}{
\begin{tabular}{ccccc}
    \toprule
	Stage & Output size & {WiGNet-Ti} & {WiGNet-S} & {WiGNet-M}
	\\
	\midrule
	Stem & $\frac{H}{4}\times\frac{W}{4}$ & {Conv$\times3$} & {Conv$\times3$} & Conv$\times3$ 
	\\
	\midrule
	{Stage 1} & {$\frac{H}{4}\times\frac{W}{4}$} 
	& $\begin{bmatrix}D=48\\E=4\\k=9\\W=8\end{bmatrix}$$\times$2 
	& $\begin{bmatrix}D=80\\E=4\\k=9\\W=8\end{bmatrix}$$\times$2 
	& $\begin{bmatrix}D=96\\E=4\\k=9\\W=8\end{bmatrix}$$\times$2 
	\\
	\midrule
	Downsample & $\frac{H}{8}\times\frac{W}{8}$ & Conv & Conv & Conv 
	\\
	\midrule
	{Stage 2} & {$\frac{H}{8}\times\frac{W}{8}$} 
	& $\begin{bmatrix}D=96\\E=4\\k=9\\W=8\end{bmatrix}$$\times$2 
	& $\begin{bmatrix}D=160\\E=4\\k=9\\W=8\end{bmatrix}$$\times$2 
	& $\begin{bmatrix}D=192\\E=4\\k=9\\W=8\end{bmatrix}$$\times$2 
	\\
	\midrule
	Downsample & $\frac{H}{16}\times\frac{W}{16}$ & Conv & Conv & Conv 
	\\
	\midrule
	{Stage 3} & {$\frac{H}{16}\times\frac{W}{16}$} 
	& $\begin{bmatrix}D=240\\E=4\\k=9\\W=8\end{bmatrix}$$\times$6 
	& $\begin{bmatrix}D=400\\E=4\\k=9\\W=8\end{bmatrix}$$\times$6 
	& $\begin{bmatrix}D=384\\E=4\\k=9\\W=8\end{bmatrix}$$\times$16 
	\\
	\midrule
	Downsample & $\frac{H}{32}\times\frac{W}{32}$ & Conv & Conv & Conv 
	\\
	\midrule
	{Stage 4} & {$\frac{H}{32}\times\frac{W}{32}$} 
	& $\begin{bmatrix}D=384\\E=4\\k=9\\W=8\end{bmatrix}$$\times$2 
	& $\begin{bmatrix}D=640\\E=4\\k=9\\W=8\end{bmatrix}$$\times$2 
	& $\begin{bmatrix}D=768\\E=4\\k=9\\W=8\end{bmatrix}$$\times$2 
	\\
	\midrule
	Head & $1\times1$ & {Pooling \& MLP} & {Pooling \& MLP} & {Pooling \& MLP} \\
	\midrule
	\multicolumn{2}{c}{Parameters (M)} & {10.8} & {27.4} & {49.7} 
	\\
	\midrule
	\multicolumn{2}{c}{MACs (B)} & {2.1} & {5.7} & {11.2} 
	\\
\bottomrule
\end{tabular}
}
        \caption{Detailed settings of WiGNet series. $D$: feature dimension, $E$: hidden dimension ratio in FFN, $k$: number of neighbors in GCN, $W$: window size, $H\times W$: input image size. `Ti' denotes tiny, `S' denotes small, and `M' denotes medium. 
	}
	\label{tab:wignn}
	
\end{table}

\noindent
\textbf{The Grapher module.}
\label{sec:grapher}
The Window-based Grapher module illustrated in Fig.~\ref{fig:arch-grapher}b is at the core of \method.
Preliminary, the feature vector generated from the Stem module (or the previous Grapher module) is processed by a fully connected layer with batch normalization.
Firstly, the \textit{Windows Partitioning} component splits the input tensor into non-overlapping windows having a fixed size of $M \times M$.
Secondly, the \textit{Dynamic Graph Convolution} component builds a graph and performs graph convolution independently for each window.
This component is in turn at the core of the Grapher and is described in detail in the following section.
Next, the \textit{Windows Reverse} component reshapes the output of the Dynamic Graph Convolution component into the original feature vector as generated by the Windows Partitioning component. 
The feature vector is then passed as input to a fully connected layer with batch normalization.
Finally, the Window-based Grapher block is completed with a skip connection.

\begin{figure*}[t]
    \centering
    \includegraphics[width=\textwidth]{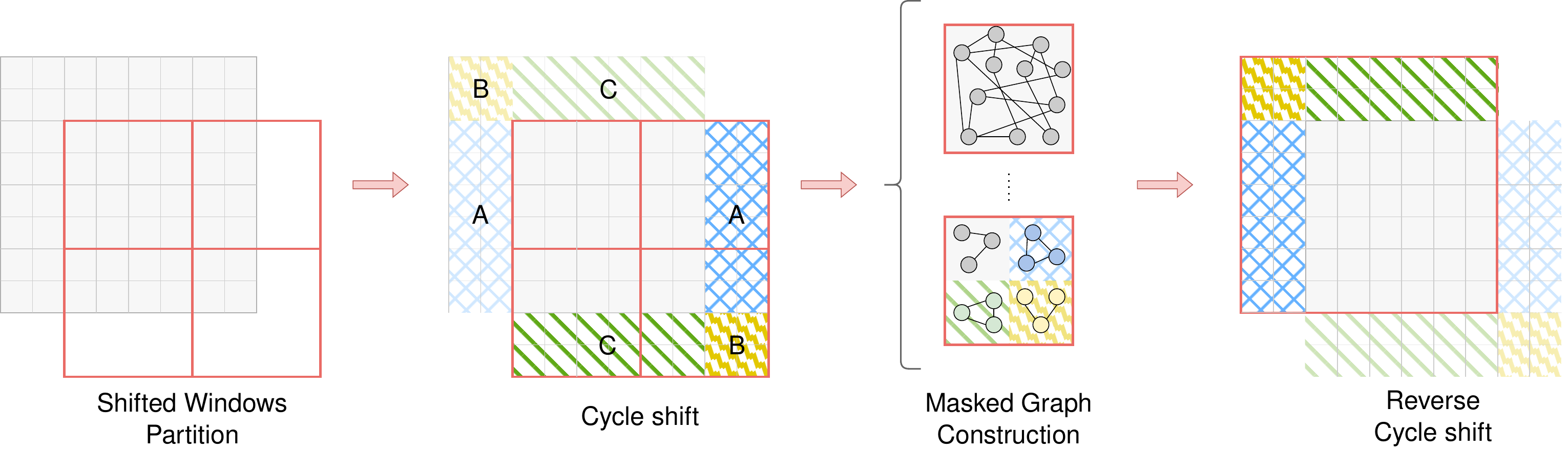}
    \caption{Overview of the cycling operation used to obtain shifted windows. The top-left part of the feature maps is copied on the bottom-right part, then the masking mechanism is used to avoid connection between non-adjacent nodes in the original feature maps.}
    \label{fig:shift-w}
\end{figure*}

\noindent
\textbf{Dynamic Graph Convolution component.}
\label{sec:dgc}
For each $w$-th window, the dynamic graph convolution component implements the $k$-NN algorithm to produce a graph $G^w=(\mathcal{V}^w,\mathcal{E}^w)$, where 
${\mathcal{E}^w\subseteq \{(i,j)\mid i,j\in \mathcal{V}^w\}}$ 
is the set of edges and $\mathcal{V}^{w}$ is the set of nodes in $G^w$.
In particular, two nodes $(i,j)$ are connected if $j \in \mathcal{N}_i^w$, where $\mathcal{N}_i^w$ is the set of $k$ nearest neighbors for the node $i$ belonging to the same window $w$.
Therefore, similarly to ViG~\cite{vgnn}, we apply the Max-Relative graph convolution proposed by Li~\etal~\cite{deepgcn} 
to update the representation of the $i$-th node in the $w$-th window as follows:
\begin{equation}
    \mathbf{x'}_{i}^{w} = \mathbf{W}_{\text{update}}\left(\mathbf{x}_{i}^{w}~\Vert~\max(\{\mathbf{x}_{j}^{w} - \mathbf{x}_{i}^{w}~\forall~j \in \mathcal{N}_i^{w}\})\right),
    \label{eq:wind-mrconv}
\end{equation}
 where $\Vert$ is the concatenation function, and $\mathbf{W}_{\text{update}}$ is a matrix of learnable parameters.

\begin{figure}[t]
    \centering
    \includegraphics[width=\columnwidth]{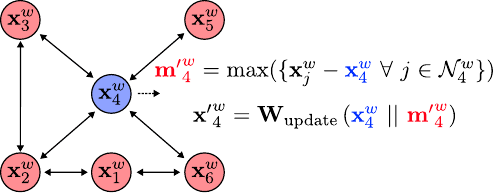}
    \caption{Illustrative example of the dynamic graph convolution of \method.}
    \label{fig:message_passing}
\end{figure}

Fig. \ref{fig:message_passing} shows an example of the computation of $\mathbf{x'}_{i}^{w}$ in \eqref{eq:wind-mrconv} for the $4$-th node, where $\mathcal{N}_4^w=\{2,3,5,6 \}$, we represent the aggregation step in \eqref{eqn:update} as $\mathbf{m'}_{4}^{w} = \max(\{\mathbf{x}_{j}^{w} - \mathbf{x}_{4}^{w}~\forall~j \in \mathcal{N}_4^{w}\})$, and the update step as $\mathbf{x'}_{4}^{w} = \mathbf{W}_{\text{update}}\left(\mathbf{x}_{4}^{w}~||~\mathbf{m'}_{4}^{w}\right)$.
For simplicity, we omit the window notation and refer to the graph convolution in \eqref{eq:wind-mrconv} as $\mathbf{X}' = \operatorname{GraphConv}(\mathbf{X})$, where $\mathbf{X} \in \mathbb{R}^{N \times F} = [\mathbf{x}_{1},\mathbf{x}_{2},\cdots\mathbf{x}_{N}]^\top$ is the set of node features.

GNNs typically include few graph convolutional layers due to the over-smoothing problem \cite{giraldo2023trade}, where features tend to be more and more similar and thus less discriminative with the network depth.
For this reason, we employ an FFN to perform feature transformations and non-linear activations after the Window-based Grapher module.
Our Grapher also has a fully connected layer before and after the dynamic graph convolution layer as in \cite{vgnn}.
Besides, the graph representations are dynamically updated with $k$-NN in every new layer as in \cite{deepgcn}.
Therefore, given an input feature $\mathbf{X} \in \mathbb{R}^{N \times F}$, the overall Window-based Grapher module transfer function can be expressed as:
\begin{equation}
    \mathbf{Y} = \sigma\left(\operatorname{GraphConv}(\mathbf{X} \mathbf{W}_{\text{in}})\right) \mathbf{W}_{\text{out}} + \mathbf{X},
    \label{eq:grapher}
\end{equation}
\noindent
where $\mathbf{W}_{\text{in}} \in \mathbb{R}^{F \times 2F}$ and $\mathbf{W}_{\text{out}} \in \mathbb{R}^{2F \times F}$ are learnable parameters of fully-connected layers that respectively increase and reduce the input feature dimension $F$, and $\sigma(\cdot)$ is a non-linear activation function.
We omit the bias terms and batch normalization in \eqref{eq:grapher} for the sake of simplicity.
The operation in \eqref{eq:grapher} results in a new feature embedding $\mathbf{Y} \in \mathbb{R}^{N \times F}$.

\noindent
\textbf{The FNN module.}
To encourage feature diversity, the output $\mathbf{Y}$ of the Grapher module is processed by the FFN module as follows.
The FFN module is implemented as a multi-layer perceptron with two fully connected layers with residual connection to $\mathbf{Y}$ given by:
\begin{equation}
    \mathbf{Z} = \sigma(\mathbf{Y}\mathbf{W}_{1})\mathbf{W}_{2} + \mathbf{Y},
    \label{eq:FFN}
\end{equation}
\noindent
where $\mathbf{W}_{1} \in \mathbb{R}^{F \times p}$ projects the input features into an $p$-dimensional space, with $p = F\times E$ and $E$ the hidden dimension ratio, and $\mathbf{W}_{2} \in \mathbb{R}^{p \times F}$ re-projects the features into the original $F$-dimensional space.
The FFN module in \eqref{eq:FFN} also contains: (i) batch normalization layer after each linear projection $\mathbf{W}_{1}$ and $\mathbf{W}_{2}$, and (ii) bias terms that we omit for the sake of simplicity.

\begin{figure*}[t]
\centering
\begin{subfigure}{.5\textwidth}
  \centering
  \includegraphics[width=0.8\linewidth]{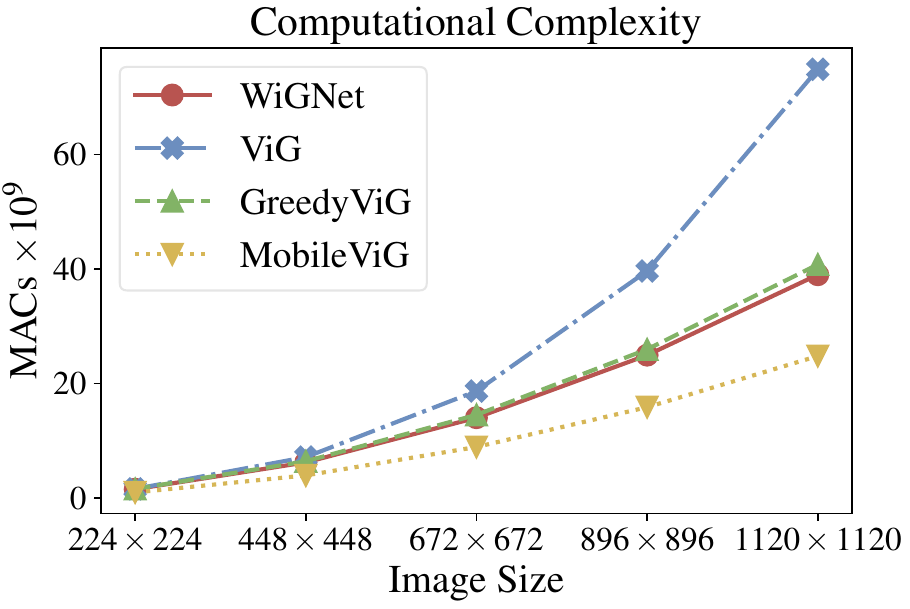}
  \caption{Computational complexity in MACs.}
  \label{fig:sub1}
\end{subfigure}%
\begin{subfigure}{.5\textwidth}
  \centering
  \includegraphics[width=0.8\linewidth]{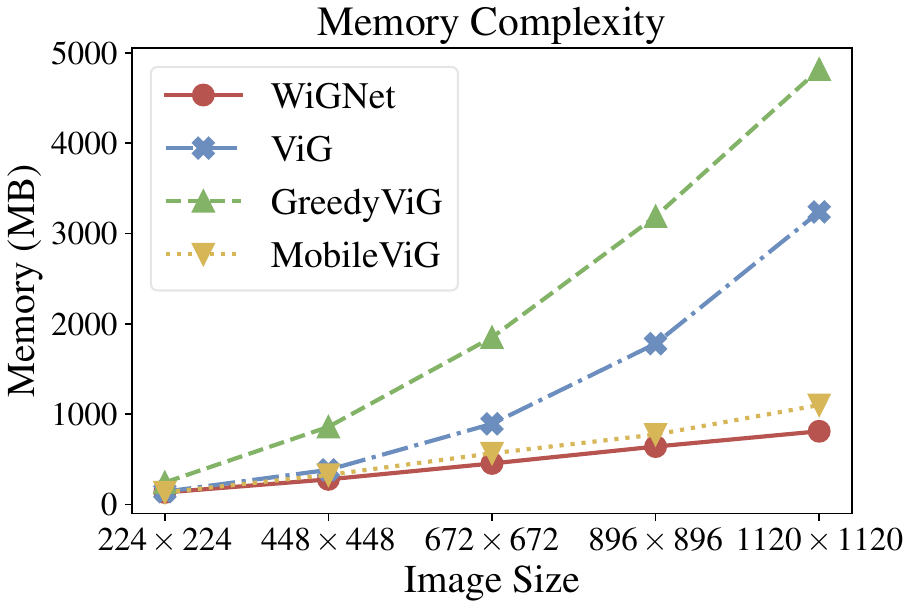}
  \caption{Memory complexity in MegaBytes (MB).}
  \label{fig:sub2}
\end{subfigure}
\caption{Computational complexity and GPU memory footprint of several vision GNN architectures and \method~in terms of MACs and MB on NVIDIA GeForce RTX 3090 GPU.}
\label{fig:computational_complexity}
\end{figure*}

\subsection{Shifted Windows.}
\label{sec:shift}
To introduce cross-window connections while maintaining the efficient computation described above, we include in \method~a shifting operator similar to the one adopted in Swin Transfomer~\cite{swin}. 
More precisely, we implement a Shifted Window-based Grapher module, where the graph construction and convolution are performed on shifted windows as illustrated in Fig.~\ref{fig:shift-w}.
To do this, we adopt a cycling operation to partition the feature map, and we use a masking mechanism to allow connection only between nodes adjacent to the feature map.
In other words, multiple sub-graphs may arise in the same window as shown in Fig.~\ref{fig:shift-w}, where different colors and texture backgrounds are used to identify the masking mechanism (\ie, connections are allowed only between nodes that fall in the same color area). 
This phenomenon results in a heterogeneous construction of the graphs, implying a considerable drop in the number of neighboring nodes in certain regions.
For instance, a node belonging to the top-left window of Fig.~\ref{fig:shift-w} will be connected to $k$ other nodes in that window out of $M \times M$ possible nodes, where $M$ is the window size. 
Instead, a node in the section \emph{B} of the bottom-right window will still be connected to \textit{k} other nodes but out of $S \times S$ possible nodes, where $S$ is the shift-size typically set as $S = \lfloor\frac{M}{2} \rfloor$.
To attempt to solve this issue, we linearly adjust the number of neighbors of each node by considering the maximum number of possible neighbors that the masking mechanism allows it to have.
In particular, given $k$, the window-size $M \times M$, and the number of possible neighbors for the node $i$ ($P_{i}$), we can use $k_{i} = k \times \frac{P_{i}}{M^{2}}$ as the number of neighbors for that node.

\subsection{Complexity Considerations.}

Although both our method and ViG use $k$-NN to create the graph, one of the major advantages of \method~is the reduction in computational complexity as the image size increases.
ViG's $k$-NN complexity, indeed, grows with the square of the number $hw$ of nodes (patches) of the whole feature map and is given by:
\begin{equation}
    \Omega \left(\text{ViG, }k\text{-NN} \right) = (hw)^2.
\end{equation}
In contrast, the windowed approach of \method~results in a complexity that grows linearly with the number of patches as follows:
\begin{equation}
    \Omega (\text{WiGNet, }k{\text{-NN}}) = \left(\frac{hw}{\vert \mathcal{V}^w \vert}\right) \vert \mathcal{V}^w \vert^2 = hw \vert \mathcal{V}^w \vert,
\end{equation}
where $\vert \mathcal{V}^w \vert$ is the number of nodes on each window $w$.
We compare Multiply–Accumulate (MACs) operations and memory footprint of ViG, \method~and two other graph-based models in Fig.~\ref{fig:computational_complexity}.

    

\section{Experiments}
\label{sec:experiments}

In this section, we first experiment with \method~over the 
ImageNet-1K~\cite{ImageNet} dataset comparing against ResNet~\cite{resnet}, Pyramid Vision Tranformer~\cite{pvt},  Swin Transformers~\cite{swin}, Poolformer~\cite{metaformer}, ViG~\cite{vgnn}, MobileViG~\cite{vgnn}, and GreedyViG~\cite{vhgnn}. For the sake of comparability, we consider $k=9$ neighbors for graph construction as in ViG~\cite{vgnn}. 
Then, once we train our model on ImageNet, we evaluate its adaptability to a new classification task with higher-resolution images.
To do this, we use our \textit{tiny} model as a pre-trained frozen backbone for facial identification on the CelebA-HQ~\cite{celebaHQ} dataset.
Finally, we perform two ablation studies on key design choices: whether or not to use shifting windows and which graph convolutional layer to adopt.

\subsection{Experimental Setup}

\noindent \textbf{Datasets.}
In image classification, the benchmark dataset ImageNet ILSVRC~2012~\cite{ImageNet} is commonly used as a standard evaluation metric.
ImageNet contains approximately $1.2$ million in training images and $50,000$ in validation images, spanning across $1,000$ categories\footnote{For information on the licensing of the ImageNet dataset, please refer to the website \url{http://www.image-net.org/download}.}.

The CelebA-HQ~\cite{celebaHQ} dataset is instead used to test the adaptability of our model in a downstream task with high-resolution images. Indeed, this dataset is a high-quality version of CelebA~\cite{celeba} that consists of $30,000$ images. 
We use this dataset to perform a facial identification of $307$ classes by rescaling the images to a resolution of $512 \times 512$. This rescaling is performed to be able to train the most memory-intensive models like ViG and GreedyViG.

\noindent \textbf{Implementation details.}
For training all WiGNet models on ImageNet we keep similar hyperparameters as ViG~\cite{vgnn}.
We adopt the commonly-used training strategy proposed in DeiT~\cite{deit} for fair comparison.
The data augmentation includes RandAugment~\cite{randaugment}, Mixup~\cite{mixup}, Cutmix~\cite{cutmix}, random erasing~\cite{erasing}.
Additionally, for \method-M we adopt the repeated augmentation~\cite{hoffer2020augment} and an Exponential Moving Average (EMA) scheme.
We implement our models using PyTroch~\cite{pytorch} and train all of them on $8$ GPUs NVIDIA GeForce RTX 3090.

Then, once we obtain our pre-trained models, we perform a transfer-learning experiment on higher-resolution images. In this context, all models are finetuned using Adam as an optimizer having a constant learning rate of $0.001$ for $30$ epochs using a Cross-Entropy loss function and a batch size of $64$, except for GreedyViG and ViG where we used a batch-size of $16$ for memory reasons.

\begin{table}[t]
	\small 
	\centering
        \setlength{\tabcolsep}{4pt}{
	\resizebox{\columnwidth}{!}{
            \begin{tabular}{lccccc}
			\toprule
			Model    &  Resolution  & Params (M) & MACs (G) & Top-1 & Top-5 \\
			\midrule
			{\color{resnet}$\spadesuit$} ResNet-18~\cite{resnet,wightman2021resnet} & 224$\times$224 & 12 & 1.8 & 70.6 & 89.7 \\[0.2em]
            \cdashline{1-6}\\[-0.8em]
			{\color{resnet}$\spadesuit$} ResNet-50~\cite{resnet,wightman2021resnet} & 224$\times$224 & 25.6 & 4.1 & 79.8 & 95.0 \\	
			{\color{resnet}$\spadesuit$} ResNet-152~\cite{resnet,wightman2021resnet} & 224$\times$224 & 60.2 & 11.5 & 81.8 & 95.9 \\
			\midrule
			{\color{transformer}$\blacklozenge$} PVT-Tiny~\cite{pvt}  & 224$\times$224 & 13.2 & 1.9 & 75.1 & - \\[0.2em]
            \cdashline{1-6}\\[-0.8em]
			{\color{transformer}$\blacklozenge$} PVT-Small~\cite{pvt}  & 224$\times$224 & 24.5 & 3.8 & 79.8 & - \\
			{\color{transformer}$\blacklozenge$} PVT-Medium~\cite{pvt}   & 224$\times$224 & 44.2 & 6.7 & 81.2 & - \\
			{\color{transformer}$\blacklozenge$} PVT-Large~\cite{pvt}  & 224$\times$224 & 61.4 & 9.8 &  81.7 & - \\
			{\color{transformer}$\blacklozenge$} Swin-T~\cite{swin}  & 224$\times$224 & 29 & 4.5 & 81.3 & 95.5 \\
			{\color{transformer}$\blacklozenge$} Swin-S~\cite{swin}  & 224$\times$224 & 50 & 8.7 & 83.0 & 96.2 \\
			\midrule
			{\color{mlp}$\blacksquare$} Poolformer-S12~\cite{metaformer}  & 224$\times$224 & 12 & 2.0 & 77.2 & 93.5 \\
			{\color{mlp}$\blacksquare$} Poolformer-S36~\cite{metaformer}  & 224$\times$224 & 31 & 5.2 & 81.4 & 95.5 \\
			{\color{mlp}$\blacksquare$} Poolformer-M48~\cite{metaformer}  & 224$\times$224 & 73 & 11.9 & 82.5 & 96.0 \\
		\midrule
            \midrule
			{\color{vig}$\clubsuit$} ViG-Ti~\cite{vgnn}& 224$\times$224 & 10.7 & 1.7 & 78.2 & 94.2 \\[0.2em]
            \cdashline{1-6}\\[-0.8em]
			{\color{vig}$\clubsuit$} ViG-S~\cite{vgnn} & 224$\times$224 & 27.3 & 4.6 & 82.1 & 96.0 \\
			{\color{vig}$\clubsuit$} ViG-M~\cite{vgnn} & 224$\times$224 & 51.7 & 8.9 & 83.1 & 96.4 \\

            \midrule

            {\color{mobilevig}$\maltese$} MobileViG-Ti~\cite{mobilevig}& 224$\times$224 & 5.2 & 0.7 & 75.7 & - \\
		{\color{mobilevig}$\maltese$} MobileViG-S~\cite{mobilevig}& 224$\times$224 & 7.2 & 1.0 & 78.2 & - \\[0.2em]
            \cdashline{1-6}\\[-0.8em]
            {\color{mobilevig}$\maltese$} MobileViG-M~\cite{mobilevig}& 224$\times$224 & 14.0 & 1.5 & 80.6 & - \\
            
            \midrule
            {\color{greedyvig}$\blacktriangledown$} GreedyViG-S~\cite{greedyvig}& 224$\times$224 & 12.0 & 1.6 & 81.1 & - \\
		{\color{greedyvig}$\blacktriangledown$} GreedyViG-M~\cite{greedyvig}& 224$\times$224 & 5.2 & 3.2 & 82.9 & - \\
			
            \midrule
            {\color{wignet}$\bigstar$} WiGNet-Ti (ours) & 224$\times$224 & 10.7 & 1.6 & 78.4 & 94.3 \\
            {\color{wignet}$\bigstar$} WiGNet-Ti (ours) & 256$\times$256 & 10.8 & 2.1 & 78.8 & 94.6 \\[0.2em]
            \cdashline{1-6}\\[-0.8em]
			{\color{wignet}$\bigstar$} WiGNet-S (ours) & 256$\times$256 & 27.4 & 5.7 & 82.0 & 95.9 \\
			{\color{wignet}$\bigstar$} WiGNet-M (ours) & 256$\times$256 & 49.7 & 11.2 & 83.0 & 96.3 \\
            
			\bottomrule
		\end{tabular}
        }
        }
        \caption{Results of WiGNet and other deep learning methods on ImageNet. {\color{resnet}$\spadesuit$} CNN, {\color{mlp}$\blacksquare$} MLP, {\color{transformer}$\blacklozenge$} Transformers, {\color{vig}$\clubsuit$} ViG, {\color{mobilevig}$\maltese$} MobileViG, {\color{greedyvig}$\blacktriangledown$} GreedyViG and {\color{wignet}$\bigstar$} WiGNet (ours).} 
        \label{tab:res-inet}
\end{table}

\subsection{Main Results}

First, we provide the classification results on ImageNet. Then, we show that our pre-trained backbone achieves a better trade-off between accuracy and complexity than other models using CelebA-HQ as a downstream task. 

\noindent \textbf{ImageNet.}
\label{sec:main_res}
Table~\ref{tab:res-inet} shows the comparison between \method~and previous state-of-the-art methods.
WiGNet outperforms or achieves competitive results against previous state-of-the-art models for similar complexity.
For instance, comparing \method~with non-graph-based models, our tiny model with $78.8$ of accuracy outperforms all previous methods for low MACs (around $2$G), and a small number of parameters (around $10$M). 
Similarly, the \method-S achieves better results than the Swin-T model with comparable MACs and than ResNet-152 with almost half the parameters.

In addition, \method~shows competitive results against the previous graph-based method under similar conditions. 
Moreover \method-Ti trained with slightly larger images ($256 \times 256$) and using a window size of $8 \times 8$, works better than the same model trained on ($224 \times 224$) images since in this case the window-size is smaller ($7 \times 7$) and thus the number of possible neighbors.



\begin{figure}[t]
    \centering
    \includegraphics[width=\linewidth]{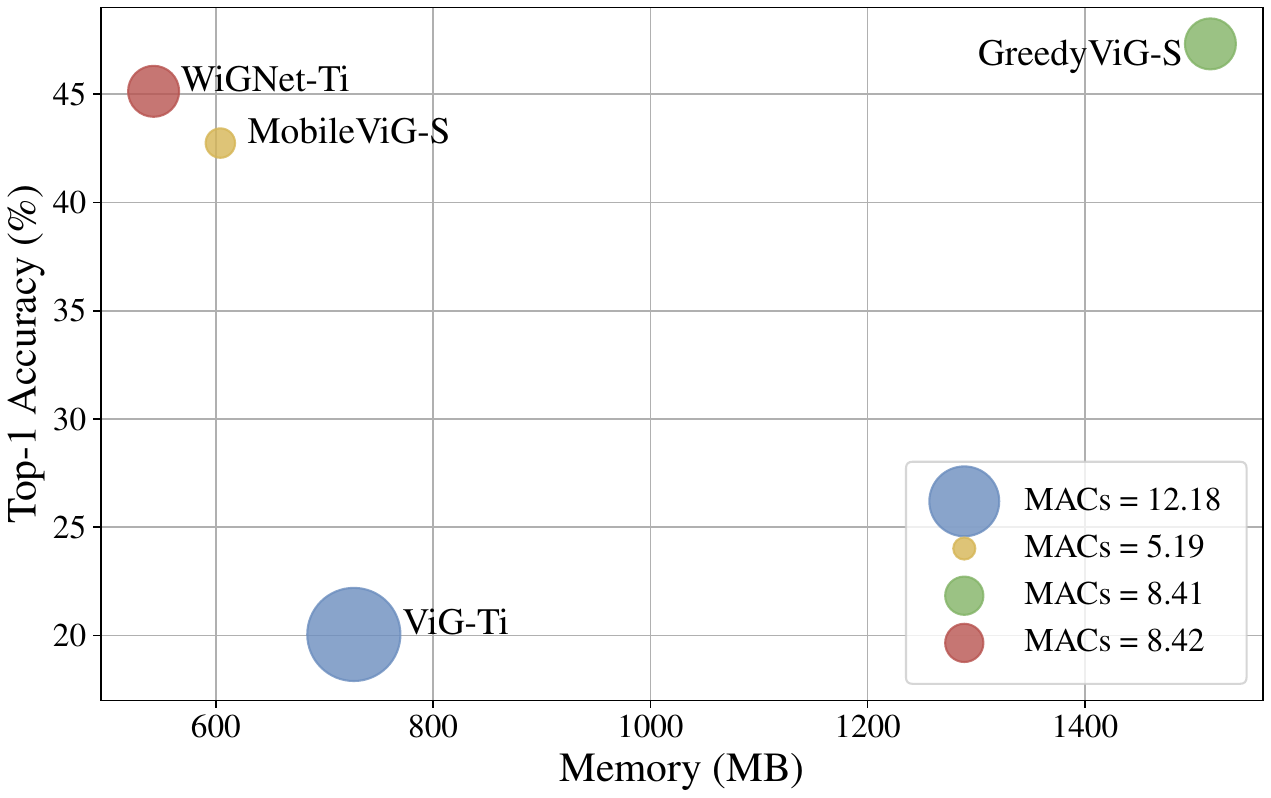}
    \caption{Comparison of Graph-based models on $512 \times 512$ resolution images. The size of the dots represents the used MACs.}
    \label{fig:plot_hq_res}
\end{figure}

\noindent \textbf{CelebA-HQ.}
To show the adaptability of \method~to new classification tasks having higher resolution images, we conduct experiments using our pre-trained model on ImageNet as a frozen backbone on the CelebA-HQ dataset~\cite{celebaHQ} as a downstream facial identity classification task. 
Particularly, a new classification layer was trained on these features keeping the rest of the architecture frozen.
Fig.~\ref{fig:plot_hq_res} shows the results obtained by our backbone compared to other graph-based models in terms of accuracy, memory usage, and MACs using $512 \times 512$ resolution images.
In this context we notice that ViG struggles to converge, while our backbone achieves the second-best result, only outperformed by GreedyViG. However, by comparing the memory footprint required for each model, we observe that \method~needs only $0.5$ GB, while for GreedyViG the occupancy is $\sim 3\times$ more. 
MobileViG, instead, is the model with the lowest MACs. Nevertheless, it occupies more memory than \method~achieving worse results.

It is clear from Fig.~\ref{fig:plot_hq_res} that \method~is the closest model to the optimal point (\ie top-left corner of the plot), achieving similar Top-1 accuracy results to GreedyViG but using significantly less memory, even compared to MobileViG.
In Fig.~\ref{fig:computational_complexity} we also analyzed the complexity of these models in terms of MACs and memory as the resolution of the input image increases. From these results we observe that \method~computation and memory requirements scale only linearly with the image size. By comparison, the memory requirements for GreedyViG (and for ViG also the complexity in terms of MACs) scales quadratically with the image size.
These results show that \method~can operate with images of high resolution using less memory than MobileViG while maintaining the complexity under control.

\subsection{Ablation Studies}



\begin{table}[t]
    \centering
    \resizebox{0.70\columnwidth}{!}{
\begin{tabular}{lccc}
            \toprule
                 Model & Resolution & Shifting & Top-1 \\\midrule
                 WiGNet-Ti & $256 \times 256$ & \XSolidBrush &  78.9\\
                 WiGNet-Ti & $256 \times 256$ & \Checkmark  &  78.8\\
                 \midrule
                 WiGNet-S & $256 \times 256$ & \XSolidBrush & 82.0\\
                 WiGNet-S & $256 \times 256$ & \Checkmark & 82.0\\
                 \midrule
                 WiGNet-M & $256 \times 256$ & \XSolidBrush & 82.9\\
                 WiGNet-M & $256 \times 256$ & \Checkmark & 83.0\\
            \bottomrule
            \end{tabular}
        }
    \caption{Ablation study in the impact of the shifting operation and the adaptive $k$-NN strategy for \method~on ImageNet.}
    \label{tab:res-abl}
\end{table}

\noindent \textbf{Shifted windows.}
Table~\ref{tab:res-abl} shows the results when the \method~uses the Shifted Window-based Grapher module 
explained in Sec.~\ref{sec:shift}.
We observe that contrary to the Swin Transformer, the shifting strategy does not bring any advantage to \method~in this context, despite the results seeming to improve slightly by increasing the model size. We hypothesize that, because of the low resolution of the images in ImageNet, is possible to independently analyze the windows and still obtain good results. Therefore, we conduct the same transfer learning experiment described in Sec.~\ref{sec:main_res} to monitor the behavior of our backbone without shifting and with higher-resolution images, ablating also on the Adaptive $k$-NN strategy.
In Tab.~\ref{tab:res-abl-hq} we observe that for larger images the shifting operator is crucial, allowing for a gain of almost 2\% points on average, and a significantly lower standard deviation. Moreover, this gain increases to 6\% when the Adaptive $k$-NN strategy is implemented.

\begin{table}[t]
    \centering
    \resizebox{0.45\textwidth}{!}{
\begin{tabular}{lcccc}
            \toprule
                 Model & Resolution & Shifting & Adaptive-$k$-NN & Top-1 \\\midrule
                 WiGNet-Ti & $512 \times 512$ & \XSolidBrush & - & 39.48 ($\pm$ 6.02)\\
                 WiGNet-Ti & $512 \times 512$ & \Checkmark & \XSolidBrush & 41.15 ($\pm$ 0.65)\\
                 WiGNet-Ti & $512 \times 512$ & \Checkmark & \Checkmark & 45.13 ($\pm$ 1.73)\\
            \bottomrule
            \end{tabular}
        }
    \caption{Ablation study in the impact of the shifting operation for higher resolution images from the CelebA-HQ dataset.}
    \label{tab:res-abl-hq}
\end{table}




\noindent \textbf{Graph Convolutional Operator.}
Finally, we conduct an ablation study with some well-known graph convolutional functions in the Grapher module, including Max-Relative GraphConv~\cite{deepgcn}, GraphSAGE~\cite{sage} and EdgeConv~\cite{edgeconv}.
Table \ref{tab:res-conv} shows the results of this experiment when the \method-Ti model is trained on ImageNet without the shifting operator, as it seems to work slightly better for the tiny size model. 
We observe that the Max-Relative graph convolution achieves competitive results with less complexity than the other operators.

\subsection{Limitations}



The main limitation of our method compared to other Graph-based approaches is the lack of global information during the feature update process. This problem is partially mitigated by the hierarchical structure of the architecture and the shifting operation, which allows to capture less local (but not global) information through cross-windows connections. However, this operation in \method~is not as straightforward as in Swin Transformers: we should dynamically adapt the $k$ value for the $k$-NN in the borders of the image to be consistent with the rest of the regions as shown in Fig.~\ref{fig:shift-w}.
By adopting this strategy, we successfully classify high-resolution images. 
Nevertheless, we believe that more global information might be useful in tasks where we need to capture long-range dependencies like, for example, image segmentation. One possible solution to this limitation is to promote connections among graphs in the same layer.
However, this solution poses practical and theoretical challenges that deserve exploration in future works.

\begin{table}[t]
	\small 
	\centering
            \resizebox{0.45\textwidth}{!}{
            \begin{tabular}{lcccc}
			\toprule
			Model    &  Graph-Conv & MACs (G) & Top-1 & Top-5 \\
			\midrule
			
            WiGNet-Ti  & Max-Relative & 2.1 & 78.9 & 94.6 \\
            WiGNet-Ti  & GraphSAGE & 2.4 & 78.4 & 94.3 \\
            WiGNet-Ti  & EdgeConv & 3.3 & 78.7 & 94.5\\

			\bottomrule
		\end{tabular}
        }
	\caption{ImageNet results using different Graph Convolutional layers. Comparison performed on the \textit{tiny} model size without the shifting operator.}\label{tab:res-conv}
\end{table}
\section{Conclusions}

This work introduced a new Windowed vision GNN (\method) for image analysis tasks.
Our model partitions the input images into windows, and therefore graphs are constructed in the local windows.
Thus, we use the Max-Relative graph convolution operation on each local window for feature updating.
\method's architecture is completed with FFN and downsampling operations.
We show in theory and practice that the computational and memory complexity of \method~scales linearly with the image size.
At the same time, for previous vision GNNs such as ViG \cite{vgnn}, the complexity grows quadratically.
This has profound implications for GNN-based vision models' applicability in tasks requiring high-resolution images.
We conducted experiments in the ImageNet-1k benchmark dataset and then we show that \method~can be successfully adopted as a pre-trained backbone for high-resolution image classification on the CelebA-HQ dataset, achieving a better trade-off between accuracy and complexity with respect to other graph-based models. 
Thus, \method~offers a strong and scalable alternative to previous deep learning models for computer vision tasks, proving suitable for working with high-resolution images.




\section*{Acknowledgements} 
This research was partially funded by Hi!PARIS Center on Data Analytics and Artificial Intelligence. This project was provided with computer and storage resources by GENCI at IDRIS thanks to the grant 2024-AD011015338 on the supercomputer Jean Zay. 

{\small
\bibliographystyle{ieee_fullname}
\bibliography{main}
}

\end{document}